\documentclass[10pt,twocolumn,letterpaper]{article}

\usepackage{cvpr}
\usepackage{times}
\usepackage{epsfig}
\usepackage{graphicx}
\usepackage{amsmath}
\usepackage{amssymb}
\usepackage{bm}
\usepackage[table]{xcolor}
\usepackage{multirow}
\usepackage{booktabs} % for pretty plots
\usepackage[english]{babel}
\usepackage{enumitem}
\usepackage[pagebackref=true,breaklinks=true,letterpaper=true,colorlinks,bookmarks=false]{hyperref}

\cvprfinalcopy % *** Uncomment this line for the final submission

\begin{document}

\setlength\abovedisplayskip{0.5em}
\setlength\belowdisplayskip{0.5em}

%%%%%%%%% TITLE
\title{Learning to Estimate 3D Human Pose and Shape from a Single Color Image}

\author{Georgios Pavlakos$^1$, Luyang Zhu$^2$, Xiaowei Zhou$^3$, Kostas Daniilidis$^1$ \\[0ex]
$^1$ University of Pennsylvania \hspace{0.1em} $^2$ Peking University \hspace{0.1em} $^3$ Zhejiang University
}

\maketitle
%\thispagestyle{empty}

%%%%%%%%% ABSTRACT
\begin{abstract}

This work addresses the problem of estimating the full body 3D human pose and shape from a single color image. This is a task where iterative optimization-based solutions have typically prevailed, while Convolutional Networks (ConvNets) have suffered because of the lack of training data and their low resolution 3D predictions. Our work aims to bridge this gap and proposes an efficient and effective direct prediction method based on ConvNets. Central part to our approach is the incorporation of a parametric statistical body shape model (SMPL) within our end-to-end framework. This allows us to get very detailed 3D mesh results, while requiring estimation only of a small number of parameters, making it friendly for direct network prediction. Interestingly, we demonstrate that these parameters can be predicted reliably only from 2D keypoints and masks. These are typical outputs of generic 2D human analysis ConvNets, allowing us to relax the massive requirement that images with 3D shape ground truth are available for training. Simultaneously, by maintaining differentiability, at training time we generate the 3D mesh from the estimated parameters and optimize explicitly for the surface using a 3D per-vertex loss. Finally, a differentiable renderer is employed to project the 3D mesh to the image, which enables further refinement of the network, by optimizing for the consistency of the projection with 2D annotations (i.e.,~2D keypoints or masks). The proposed approach outperforms previous baselines on this task and offers an attractive solution for direct prediction of 3D shape from a single color image.

\end{abstract}

%%%%%%%%% BODY TEXT
\section{Introduction}
Estimating the full body 3D pose and shape of humans from images has been a challenging goal of computer vision going all the way back to the work of Hogg~\cite{hogg1983model}. The inherent ambiguity of the problem has forced the researchers to use monocular image sequences for inference~\cite{xu2017monoperfcap,alldieck2017optical}, employ multiple camera views~\cite{rhodin2016general,huang2017towards}, or even explore alternative sensors, like Kinect~\cite{weiss2011home} or IMUs~\cite{von2017sparse}. In these settings, the body shape reconstruction results are remarkable. However, estimating 3D pose and shape from single color images remains the ultimate goal for 3D human analysis.

Considering the particularly challenging nature of such a problem, the literature remains undeniably sparse. Most approaches rely on iterative optimization, attempting to estimate a full body 3D shape that is consistent with 2D image observations, like silhouettes, edges, shading, or 2D keypoints~\cite{sigal2008combined,guan2009estimating}. Despite the significant runtime required to solve the complicated optimization problem, the common failures because of local minima, and the error-prone reliance on ambiguous 2D cues, optimization-based solutions remain the leading paradigm for this problem~\cite{lassner2017unite,bogo2016keep}. Even the emergence of deep learning has not changed significantly the landscape. ConvNets did not seem as a viable candidate for this problem because they require a huge amount of training data and they are infamous for their low resolution 3D predictions~\cite{riegler2016octnet,tatarchenko2017octree}. The goal of our work is to demonstrate that ConvNets can indeed offer an attractive solution for this problem, by proposing an efficient and effective direct prediction approach, which is competitive and even outperforms iterative optimization methods.

\begin{figure*}[t]
	% minipage mit (Blind-)Text
	  \centering
	  \includegraphics[width=0.95\linewidth,trim={0cm 11cm 0cm 0cm},clip]{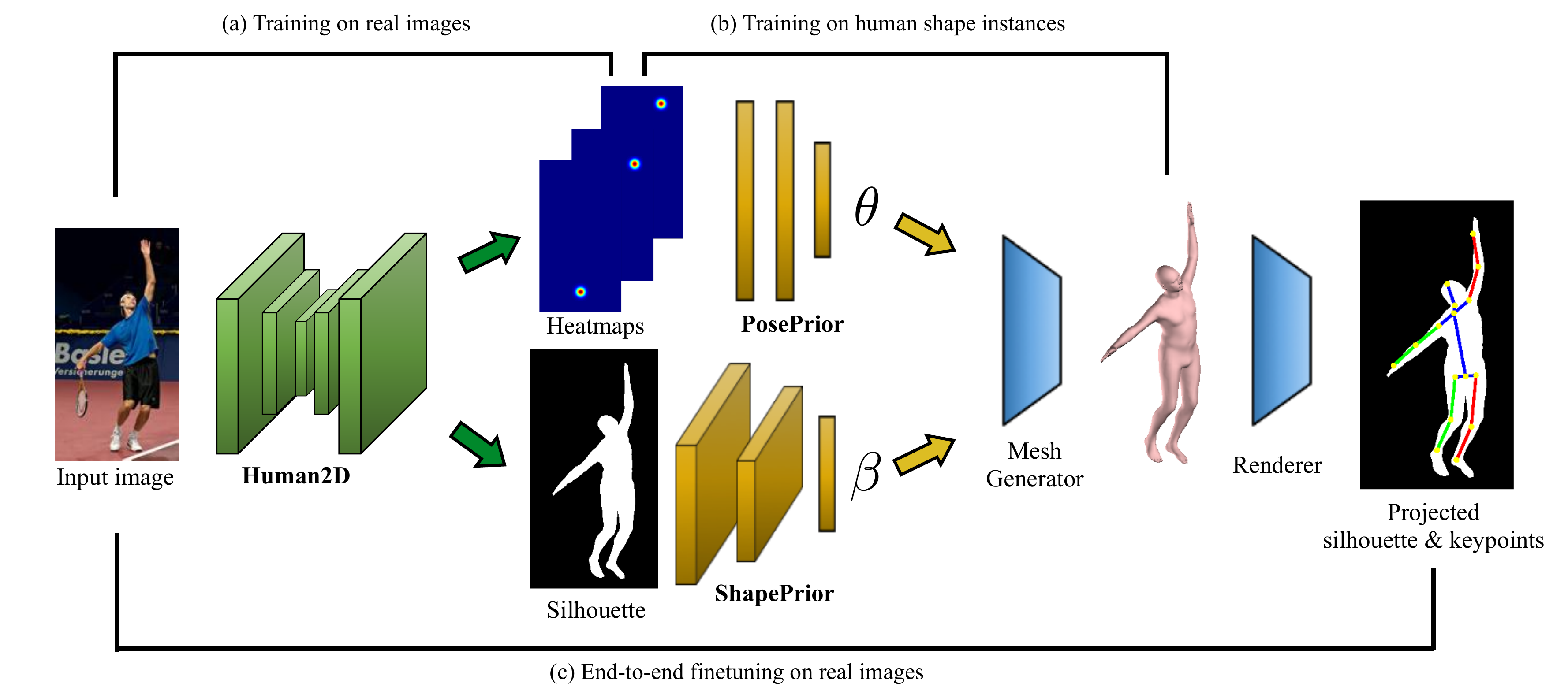}
    \caption{
Schematic representation of our framework. (a) An initial ConvNet, {\em Human2D}, predicts 2D heatmaps and masks from a single color image, using 2D pose data~\cite{johnsonclustered,andriluka2014mpii} for training. (b) Two networks estimate the parameters of the statistical model SMPL~\cite{loper2015smpl}, using instances of the parametric model for training. The {\em PosePrior} estimates pose parameters ($\theta$) from keypoints, and the {\em ShapePrior} estimates shape parameters ($\beta$) from silhouettes. (c) The framework can be finetuned end-to-end without requiring images with 3D shape ground truth, by projecting the full body 3D mesh to the image and optimizing for the consistency of the projection with 2D annotations (keypoints and masks). The blue parts (Mesh Generator and Renderer) indicate components without learnable parameters.
}
 \label{fig:pipeline}
\end{figure*}

To make this feasible, a critical design choice for our approach is the incorporation of a parametric statistical body shape model (SMPL~\cite{loper2015smpl}) within our end-to-end framework, presented in Figure~\ref{fig:pipeline}. The advantage of such a representation is that we can generate high quality 3D meshes in the form of 6890 vertices while estimating only a small number of parameters, i.e., 72 for pose and 10 for shape. This low-dimensional parameterization makes the model friendly for direct network prediction. In fact, this prediction is feasible and accurate by using only 2D keypoints and silhouettes as input. This allows us to relax the limiting assumption that natural images with 3D shape ground truth are available for training. In contrast, we can leverage the available 2D image annotations (e.g.,~\cite{johnsonclustered,andriluka2014mpii}) to train for image-to-2D inference, while using instances of the parametric model to train for 2D-to-3D shape inference. Simultaneously, another major advantage of employing this parametric model is that its structure allows us to generate the estimated 3D mesh at training time and optimize directly for the surface, by using a 3D per-vertex loss. This loss has better correlation with the vertex-to-vertex 3D error that is typically used for evaluation and improves training compared to naive parameter regression. Finally, we propose to employ a differentiable renderer to project the generated 3D mesh back to the 2D image. This enables end-to-end finetuning of the network by optimizing for the consistency of the projection with annotated 2D observations, i.e., 2D keypoints and masks. The complete framework offers a modular direct prediction solution to the problem of 3D human pose and shape estimation from a single color image and outperforms previous approaches on the relevant benchmarks.

Our main contributions can be summarized as follows:
\begin{itemize}
\itemsep0em
\item an end-to-end framework for 3D human pose and shape estimation from a single color image.
\item incorporation of a parametric statistical shape model, SMPL, within the end-to-end framework, enabling:
	\begin{itemize}[leftmargin=.1in]
	\itemsep0em
	\item prediction of the SMPL model parameters from ConvNet-estimated 2D keypoints and masks to avoid training on synthetic image examples.
	\item generation of the 3D body mesh at training time and supervision based on the 3D shape consistency.
	\item use of a differentiable renderer for 3D mesh projection and refinement of the network with supervision based on the consistency with 2D annotations.
	\end{itemize}
\item superior performance compared to previous approaches for 3D human pose and shape estimation at significantly faster running time.
\end{itemize}

%-------------------------------------------------------------------------
\section{Related work}
\noindent
\textbf{3D human pose estimation}:
In order to estimate a convincing 3D reconstruction of the human body, it is crucial to get an accurate prediction of the 3D pose of the person. Many recent works follow the end-to-end paradigm~\cite{tome2017lifting,rogez2017lcr,sun2017compositional,tekin2017learning,zhou2017towards}, using images as input to predict 3D joint locations~\cite{li20143d,tekin2016structured,popa2017deep,mehta2017monocular}, regress 3D heatmaps~\cite{pavlakos2016coarse}, or classify the image in a particular pose class~\cite{rogez2016mocap,rogez2017lcr}. Unfortunately, an important constraint is that most of these ConvNets require images with 3D pose ground truth for training, limiting the available training data sources. Other approaches commit to the 2D pose estimates provided by state-of-the-art ConvNets and focus on the 3D pose reconstruction~\cite{moreno20163d,zhou2016sparseness}, recover 3D pose exemplars~\cite{chen20163d}, or produce multiple 3D pose candidates consistent with the 2D pose~\cite{jahangiri2017generating}. Notably, Martinez~\etal~\cite{martinez2017simple} demonstrate state-of-the-art results using a simple multi-layer perceptron which regresses the 3D joint locations from 2D pose input. Our goal is significantly different from the aforementioned works, since instead of a rough stickman-like figure, we estimate the whole surface geometry of the human body.

\noindent
\textbf{Human shape estimation}:
Concurrently with advances in 3D human pose, a different set of works addressed the problem of human shape estimation. In this case, given a single image, most methods attempt to estimate the parameters of a statistical body shape model like SCAPE~\cite{anguelov2005scape} or SMPL~\cite{loper2015smpl}. The input is usually silhouettes, while regression forests~\cite{dibra2016hs} and ConvNets~\cite{dibra2016shape,dibra2017human} have been proposed for the prediction. Knowledge of human shape is useful for biometric applications, however we argue that for 3D perception the potential and the challenges are significantly greater when pose and shape are inferred jointly.

\noindent
\textbf{Joint 3D human pose and shape estimation}:
Despite individual advances in pose and shape prediction, their joint estimation makes the task significantly harder. This has consistently fostered research in non single image scenarios, for more robust results. Xu~\etal~\cite{xu2017monoperfcap} propose a pipeline for full performance capture from monocular video assuming knowledge of the shape mesh for the observed subject. Alldieck~\etal~\cite{alldieck2017optical} estimate pose and shape jointly from monocular video relying on optical flow cues. Rhodin~\etal~\cite{rhodin2016general} and Huang~\etal~\cite{huang2017towards} use images from multiple calibrated cameras and rely on keypoint detections, silhouettes and temporal consistency to recover a reconstruction of the body. An alternative setting is proposed by Weiss~\etal~\cite{weiss2011home} making use of the depth modality of the Kinect sensor to tackle the same problem. In the same spirit of exploring different sensors, von Marcard~\etal~\cite{von2017sparse} use a sparse set of IMUs on the subject to recover pose and shape jointly.

\noindent
\textbf{3D human pose and shape from a single color image}:
In the most challenging case of using only a single color image as input, the work of Sigal~\etal~\cite{sigal2008combined} is among the first to estimate high quality 3D shape estimates, by fitting the parametric model SCAPE~\cite{anguelov2005scape} to ground truth image silhouettes. Guan~\etal~\cite{guan2009estimating} use silhouettes, edges and shading as cues during the fitting process, but still require initialization through a user specified 2D skeleton. A fully automatic approach was proposed very recently by Bogo~\etal~\cite{bogo2016keep}. They use 2D keypoint detections from a 2D pose ConvNet~\cite{pishchulin2016deepcut} and fit the parametric model SMPL~\cite{loper2015smpl} to these 2D locations. Their 3D pose results are very accurate, but shape remains highly underconstrained. To improve upon this, Lassner~\etal~\cite{lassner2017unite} extends the fitting using silhouettes provided by a segmentation ConvNet. The common theme of these works is that they pose an optimization problem and attempt to fit a body model to a set of 2D observations. The drawback though is that solving this iterative optimization problem is very slow, it can easily fail because of local minima, and it relies a lot on error-prone 2D observations.

Alternatively, direct prediction approaches estimate 3D pose and shape in a discriminative way, without explicitly optimizing a specific objective during inference. Relevant to this paradigm is the work of Lassner~\etal~\cite{lassner2017unite}, where a ConvNet detects 91 landmarks of the human body and then a random forest estimates the 3D body and shape from these detections. However, to train for these landmarks, they still require alignment of body shapes with images. In contrast, we demonstrate that only a much smaller set of annotations are critical for the reconstruction, i.e., 2D joints and masks, which can be provided by human annotators and are abundant for in-the-wild images~\cite{johnsonclustered,andriluka2014mpii,lin2014microsoft}, while we also incorporate everything within a unified end-to-end framework. Concurrently, Tan~\etal~\cite{tan2017indirect} use an encoder-decoder ConvNet, where the decoder is trained to predict the silhouette corresponding to SMPL parameters. We differ to them by identifying that from these parameters we can analytically generate the body mesh and project it to the image in a differentiable way (as in~\cite{tewari2017mofa} for face models), avoiding half a million of extra learnable weights. Instead, we focus our computational and learning effort in the image to 3D shape part of the framework. Our work is also related to the concurrent work of Tung~\etal~\cite{tung2017self}, however our framework can be trained from scratch instead of relying on synthetic image data for pretraining, and we demonstrate state-of-the-art results for model-based 3D pose and shape prediction.

\section{Human body shape models}
Statistical body shape models, like SCAPE~\cite{anguelov2005scape} or SMPL~\cite{loper2015smpl}, are powerful tools, which provide significant opportunities for an end-to-end framework. One of the important advantages is their low-dimensional parameter space, which is very suitable for direct network prediction. With this parameter representation, we can keep the output prediction space small, compared to voxelized or point cloud representations. Simultaneously, the low dimensional prediction does not sacrifice the quality of the output, since we can still generate high quality 3D meshes from the estimated parameters. Furthermore, from a learning perspective, we bypass the problem of learning the statistics of the human body, and devote the network capacity at the inference of the model parameters from image evidence. In contrast, approaches without the aid of a model put additional burden on the learning side, which often leads to embarrassing prediction errors (e.g.,~failing to reconstruct limbs under occlusion, missing body details, etc). Moreover, most models offer a convenient disentanglement of pose and shape which is useful to independently focus on the factors that affect each one of the two. Last but certainly not least for end-to-end approaches, the function which generates the 3D mesh from parameter inputs is differentiable, making the models compatible with current end-to-end pipelines.

In this work, we employ the more recent SMPL model, introduced by Loper~\etal~\cite{loper2015smpl}. We provide the essential notation here, and we refer the reader to~\cite{loper2015smpl} for more details. SMPL defines a function $\mathcal{M}(\bm{\beta},\bm{\theta};\Phi)$, where $\bm{\beta}$ are the shape parameters, $\bm{\theta}$ are the pose parameters and $\Phi$ are fixed parameters of the model. The direct output of this function is a body mesh $\bm{P} \in \mathbb{R}^{N \times 3}$ with $N = 6890$ vertices $P_i \in \mathbb{R}^3$. The shape of the model uses a linear combination of a low number of principal body shapes which are learned from a large dataset of body scans~\cite{robinette2002civilian}. The {\em shape parameters} $\bm{\beta}$ are the linear coefficients of these base shapes. The pose of the body is defined through a skeleton rig with 23 joints. The {\em pose parameters} $\bm{\theta}$ are expressed in the axis angle representation and define the relative rotation between parts of the skeleton. In total, 72 parameters define the pose (3 for each of the 23 joints, plus 3 for the global rotation). Given the rest pose shape retrieved by the shape parameters $\bm{\beta}$, SMPL defines pose-dependent deformations and uses the pose parameters $\bm{\theta}$ to produce the final output mesh. Conveniently, the {\em body joints} $\bm{J}$ are a linear combination of a sparse set of mesh vertices, making joints a direct outcome of the estimated body mesh.

\section{Technical approach}
The conventional ConvNet-based approach for our task would be to acquire a large amount of color images with 3D shape ground truth and train the network with these input-output pairs. However, except for small-scale datasets~\cite{lassner2017unite} or synthetically generated image examples~\cite{varol2017learning} this type of data is typically unavailable. Therefore, to deal with this task, we need to rethink the typical pipeline. Our main goal is to leverage all the resources we have available and use our insights for the problem to build an effective framework. As a first step, from findings of prior work, we identify that 3D pose can be estimated reliably from 2D pose estimates~\cite{bogo2016keep,martinez2017simple}, while the shape can be inferred from silhouette measurements~\cite{dibra2016shape,dibra2017human}. This observation conveniently decomposes the problem in a) estimation of keypoints and masks from color images and, b) prediction of 3D pose and shape from the 2D evidence. The advantage of this practice is that the framework can be trained without requiring images with 3D shape ground truth.

\subsection{Keypoints and silhouette prediction}
The first step of our framework focuses on 2D keypoint and silhouette estimation. This part is motivated by the availability of large-scale benchmarks~\cite{johnsonclustered,andriluka2014mpii,lin2014microsoft} with 2D joints and mask annotations. Considering the volume and the variability of this data, we leverage it to train a ConvNet for 2D pose and silhouette prediction, that is particularly reliable under various imaging conditions and poses.

In the past, two individual ConvNets have been used to provide 2D keypoints and masks~\cite{huang2017towards,lassner2017unite}. In contrast, for a more elegant solution, we train a single ConvNet, which we denote as {\em Human2D}, that generates two outputs, one for keypoints and one for silhouettes. {\em Human2D} follows the Stacked Hourglass design~\cite{newell2016stacked}, using two hourglasses, which was found to be a good trade-off between accuracy and running time. The keypoint output is in the form of heatmaps~\cite{tompson2014joint,pfister2015flowing}, where an MSE loss, $\mathcal{L}_{hm}$, between the ground truth and the predicted heatmaps is used for supervision. The silhouette output has two channels (body and background) and is supervised using a pixelwise binary cross entropy loss, $\mathcal{L}_{sil}$. For training, we combine the two losses: $\mathcal{L}_{hg} = \lambda \mathcal{L}_{hm} + \mathcal{L}_{sil}$, where $\lambda = 100$. This ConvNet falls under the multi-task learning paradigm~\cite{popa2017deep}. Through sharing, the two tasks might benefit each other, but multi-task learning can also pose certain challenges (e.g.,~appropriate weighting of the losses), as Kokkinos identifies~\cite{kokkinos2016ubernet}.

\subsection{3D pose and shape prediction}
The second step is significantly more challenging, requiring estimation of the full body 3D pose and shape from 2D keypoints and silhouettes. Silhouettes and/or keypoints have been used extensively for 3D model fitting through iterative optimization~\cite{balan2007detailed,bogo2016keep,lassner2017unite}. Here, we demonstrate that this mapping can also be learned from data while it is possible to get a reliable prediction in a single estimation step.

For this mapping, we train two network components: (a) the {\em PosePrior}, which uses 2D keypoint locations as input together with the confidence of the detections (realised by the maximum value of each heatmap) and estimates the pose coefficients $\bm{\theta}$, and (b) the {\em ShapePrior}, which uses the silhouette as input and estimates the shape coefficients $\bm{\beta}$. In general, the silhouette can be helpful for 3D pose inference~\cite{balan2007detailed} and vice versa~\cite{bogo2016keep}. However, empirically we discovered this disentanglement to provide more stable and accurate 3D predictions, while it also leads to a more modular pipeline (e.g. updating only the {\em PosePrior}, without retraining the whole network). Regarding the architecture, the {\em PosePrior} uses two bilinear units~\cite{martinez2017simple}, where the input is the 2D keypoint locations and the maximum responses from each heatmap, and the output is the 72 SMPL pose parameters $\bm{\theta}$. The ShapePrior uses a simple architecture with five $3 \times 3$ convolutional layers, each one followed by max-pooling, and an additional bilinear unit at the end with 10 outputs, corresponding to the SMPL shape parameters $\bm{\beta}$.

\begin{figure}[t]
	  \centering
	  \includegraphics[width=1\linewidth,trim={0cm 15cm 8cm 0cm},clip]{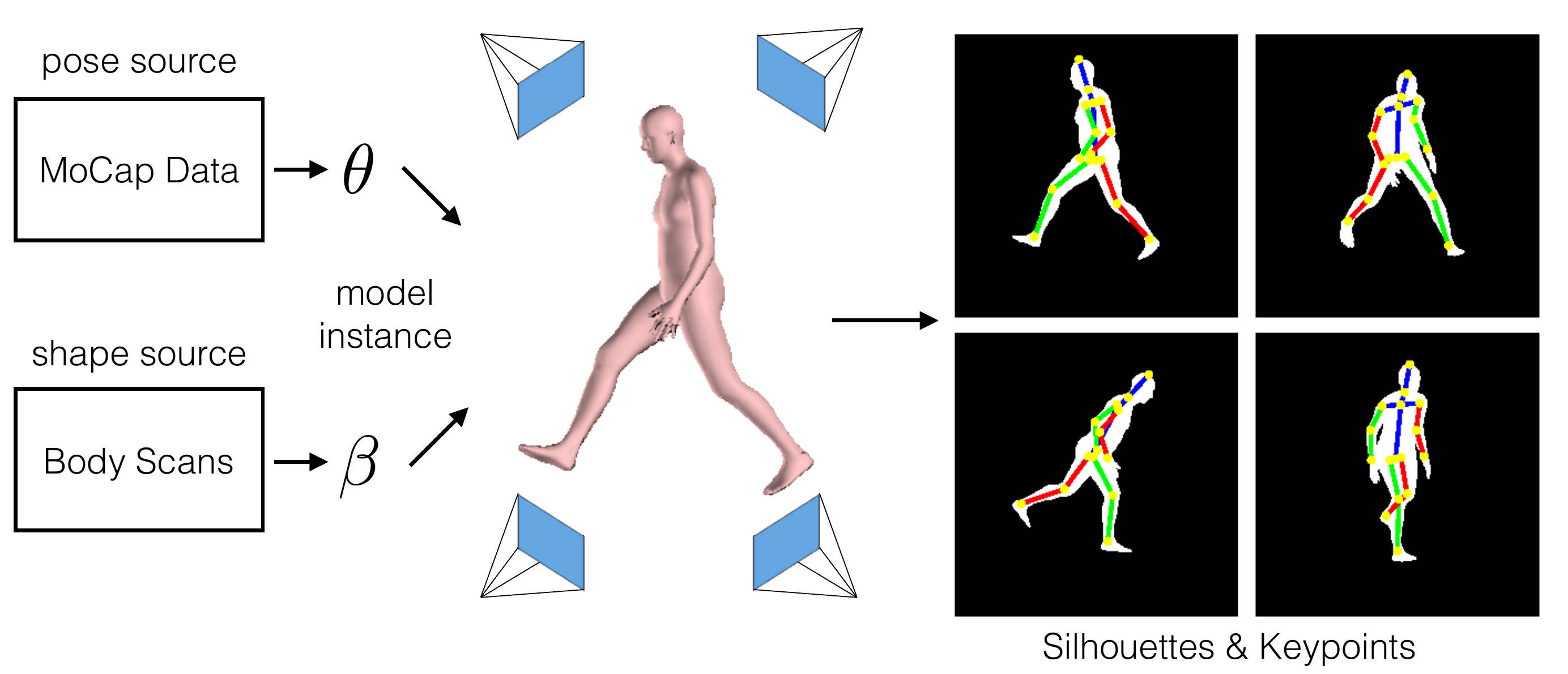}
    \caption{
We aim to learn the mapping from silhouettes and keypoints to model parameters, so we can synthesize body model instances and project them to the image plane to simulate the network input. We only require a source to sample pose parameters, and a source to sample body shape parameters. Projections from different viewpoints can also be employed for data augmentation.
}
 \label{fig:data}
\end{figure}

The form of the input (2D keypoints and masks) and the output (shape and pose parameters) allows us to produce large amount of training data by generating instances of the SMPL model with different 3D pose and shape (Figure~\ref{fig:data}). In fact, we can leverage MoCap data (e.g.,~\cite{cmuMocap,ionescu2014human}) to sample 3D poses, and body scans (e.g.,~\cite{robinette2002civilian}) to sample body shapes. For the input, we only need to project the 3D model to the image plane (possibly from different viewpoints), and compute silhouettes and 2D keypoint locations to generate input-output pairs for training. This data generation is feasible, exactly because we used an intermediate silhouette and keypoints representation. In contrast, attempting to learn a mapping directly from color images would require generation of synthetic image examples~\cite{varol2017learning}, which typically do not reach the variability of in-the-wild images.

In the previous paragraphs, we deliberately avoided discussing the supervision of the {\em Priors} networks. Past works~\cite{lassner2017unite,tan2017indirect} have examined supervision schemes using a typical $\mathcal{L}_2$ loss between the predicted and ground truth parameters. One shortcoming of this naive parameter regression approach, is that different parameters might have effects of different scale on the final reconstruction (e.g.,~the global body rotation is much more crucial than the local rotation of the hand with respect to the wrist). To avoid hand-selecting or tuning the supervision for each parameter, we aim for a more global solution. Our approach entails the generation of the full body mesh at training time, where we optimize explicitly for the predicted surface by applying a 3D per-vertex loss. Since the function $\mathcal{M}(\bm{\beta},\bm{\theta};\Phi)$ is differentiable, we can backpropagate through it and handle this mesh generator as a typical layer of our network, without any learnable parameters. Given the predicted mesh vertices $\hat{P}_i$ and the corresponding groundturth vertices $P_i$, we can supervise the network with a {\em 3D per-vertex loss}:
\begin{eqnarray}
\mathcal{L_{\mathcal{M}}} = \sum_{i=1}^N \| \hat{P}_i - P_i \|_2^2,
\end{eqnarray}
which considers all the vertices equally and has better correlation with the 3D per-vertex error which is usually employed for evaluation. Alternatively, if the focus is mainly on 3D pose, we can also supervise the network considering only the $M$ relevant 3D joints $J_i$, which are trivially exposed by the model as a sparse linear combination of the mesh vertices. In this case, denoting with $\hat{J}_i$ the estimated joints, the corresponding loss can be expressed as:
\begin{eqnarray}
\label{eqn:pose3D}
\mathcal{L_{\mathcal{J}}} = \sum_{i=1}^M \| \hat{J}_i - J_i \|_2^2.
\end{eqnarray}
Empirically, we found that the best training strategy is to initially get a reasonable initialization for the network parameters using an $\mathcal{L}_2$ parameter loss, and then activate also the vertex loss $\mathcal{L_{\mathcal{M}}}$ (or the joints loss $\mathcal{L_{\mathcal{J}}}$ if the focus is on pose only), to train a better model.

\subsection{Differentiable renderer}
Our previous analysis relaxed the assumption that images with 3D shape ground truth are available for training and relied on geometric 3D data (MoCap and body scans). In some cases though, even this type of data might be unavailable. For example, LSP~\cite{johnsonclustered} has gymnastics or parkour poses which are not represented in typical MoCap. Luckily, our generated 3D mesh has potential to leverage these 2D annotations for training purposes.

To close the loop, our complete approach includes an additional step that projects the 3D mesh to the image and examines consistency with 2D annotations. In concurrent work, a decoder-type network was used to learn the mapping from SMPL parameters to silhouettes~\cite{tan2017indirect}. However, here we identify that this mapping is known and involves the projection of the 3D mesh to the image, which can be expressed in a differentiable way, without the need to train a network with learnable weights. More specifically, for our implementation, we employ an approximately differentiable renderer, OpenDR~\cite{loper2014opendr}, which projects the mesh and the 3D joints to the image space, and enables backpropagation. The projection operation $\Pi$ gives rise to: (a) the silhouette $\Pi(\hat{\bm{P}}) = \hat{S}$, which is represented as a $64 \times 64$ binary image, and (b) the projected 2D joints $\Pi(\hat{\bm{J}}) = \hat{\bm{W}} \in \mathbb{R}^{M \times 2}$. In this case, the supervision comes from the comparison of these projections with the annotated silhouettes $S$, and the 2D keypoints $\bm{W}$, using $\mathcal{L}_2$ losses:
\begin{eqnarray}
\mathcal{L}_{\Pi} = \mu \sum_i^M \| \hat{W}_i - W_i \|_2^2 + \| \hat{S} - S \|_2^2,
\end{eqnarray}
where $\mu=10$. The goal of this type of supervision is twofold: (a) it can be employed for end-to-end refinement of the network, using only images with 2D keypoints and/or masks for training, and (b) it can be useful to mildly adapt a generic pose or shape prior to a new setting (e.g.,~new dataset), where only 2D annotations are available.

\section{Empirical evaluation}
This section focuses on the empirical evaluation of the proposed approach. First, we present the benchmarks that we employed for quantitative and qualitative evaluation. Then, we provide some essential implementation details of the approach. Finally, quantitative and qualitative results are presented on the selected datasets.

\subsection{Datasets}
For the empirical evaluation, we employed two recent benchmarks that provide color images with 3D body shape ground truth, the UP-3D dataset~\cite{lassner2017unite} and the SURREAL dataset~\cite{varol2017learning}. Additionally, we used the Human3.6M~\cite{ionescu2014human} dataset for further evaluation of the 3D pose accuracy.

\noindent
\textbf{UP-3D}: It is a recent dataset that collects color images from 2D human pose benchmarks, like LSP~\cite{johnsonclustered} and MPII~\cite{andriluka2014mpii} and uses an extended version of SMPLify~\cite{bogo2016keep} to provide 3D human shape candidates. The candidates were evaluated by human annotators to select only the images with good 3D shape fits. It comprises 8515 images, where 7818 are used for training and 1389 for testing. We report results on this test set, while we also consider subsets, based on the original dataset (LSP, MPII, or FashionPose) of the UP-3D images. Finally, we examine a reduced test set of 139 images, selected by Tan~\etal~\cite{tan2017indirect} aiming to limit the range for the global rotation. We report results using the mean per-vertex error, between predicted and ground truth shape.

\noindent
\textbf{SURREAL}: It is a recent dataset which provides synthetic image examples with 3D shape ground truth. The dataset draws poses from MoCap~\cite{cmuMocap,ionescu2014human} and body shapes from body scans~\cite{robinette2002civilian} to generate valid SMPL instances for each image. The synthetic images are not very realistic, but the accurate ground truth, makes it a useful benchmark for evaluation. We report results on the Human3.6M part of the dataset, considering all test videos and keeping every fifth frame of each video to avoid excessive redundancy in the data. Results are reported using the mean per-vertex error.

\noindent
\textbf{Human3.6M}: It is a large-scale indoor dataset that contains multiple subjects performing typical actions like ``Eating'' and ``Walking''. We follow the protocol of Bogo~\etal~\cite{bogo2016keep} using all videos of subjects S9 and S11 from `cam3' for evaluation. The original videos are downsampled from 50fps to 10fps to remove redundancy as is done in~\cite{lassner2017unite}. Results are reported using the reconstruction error.

\subsection{Implementation details}

The {\em Human2D} network is trained on MPII~\cite{andriluka2014mpii}, LSP~\cite{johnsonclustered} 
and LSP-extended~\cite{johnson2011learning} data, using the silhouettes from Lassner~\etal~\cite{lassner2017unite}. We use a batch size of 4, learning rate set to 3e-4, and rmsprop for the optimization. Augmentation for rotation ($\pm 30^{\circ}$), scale (0.75-1.25) and flipping (left-right) is used. The training lasts for 1.2M iterations.

For the {\em Priors} networks, we train with a batch size of 256, learning rate set to 3e-4, and using rmsprop for the optimization. Initially, the networks are trained for 40k iterations using an $\mathcal{L}_2$ parameter loss, and then for 60k more iterations using also $\mathcal{L}_{\mathcal{M}}$ (or $\mathcal{L}_{\mathcal{J}}$ if we focus on pose only) weighted equally with the parameter loss.

The end-to-end refinement with the reprojection loss lasts for 2k iterations with a batch size of 4, learning rate set to 8e-5, and using rmsprop for the optimization. To improve training robustness, the end-to-end updates are alternated with individual updates of the {\it Human2D} and the {\em Priors} networks (as described in the previous two paragraphs). This helps the individual components to maintain their original purpose, while we are also leveraging the strength of end-to-end training to integrate them together.

\subsection{Component evaluation}
In this section, we evaluate the components of our approach, using the UP-3D dataset. We train two different versions of our system, where for {\em Priors} we leverage data either from UP-3D (provided by Lassner~\etal~\cite{lassner2017unite}), or from CMU MoCap (provided by Varol~\etal~\cite{varol2017learning}). The {\em Human2D} network remains the same in both cases.

\begin{table}
\centering
\small
\hspace{-3mm}
\tabcolsep=0.75mm
\begin{tabular}{@{}lcc@{}}
\toprule
& \multicolumn{2}{c}{Avg error} \\
\midrule
Data source for {\em Priors} & UP-3D & CMU \\
\midrule
Parameter loss (axis-angle) & 514.9 & 589.9 \\
\midrule
Parameter loss (rot matrix) & 140.7 & 152.2 \\
\quad + Per-vertex loss & 120.7 & 142.0 \\
\quad \quad + Reprojection finetuning & 117.7 & 135.5 \\
\bottomrule
\end{tabular}
\vspace{3mm}
\caption{Ablative study on UP-3D, comparing the different supervision forms on the same architecture. The numbers are mean per-vertex errors (mm). Two versions of the {\em Priors} networks are used, trained with data from UP-3D~\cite{lassner2017unite} and CMU~\cite{varol2017learning} respectively. All networks are trained for the same number of iterations.
}
\label{tab:up3d_ablative}
\end{table}

\begin{figure}[t]
	  \centering
	   \includegraphics[width=1\linewidth,trim={0cm 10.5cm 10.3cm 0cm},clip]{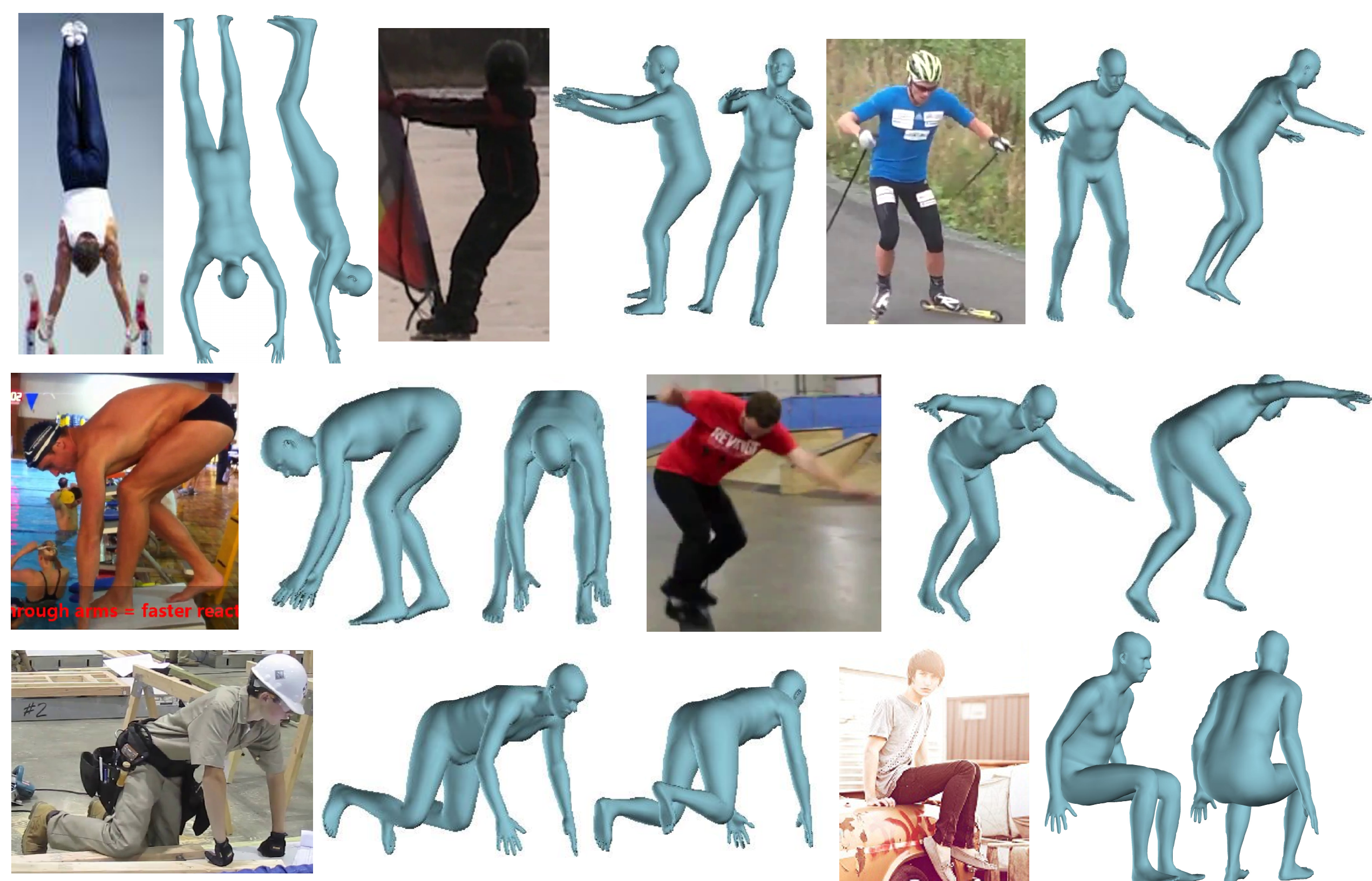}
    \caption{
Successful 3D pose and shape predictions of our approach on challenging examples of UP-3D.
}
 \label{fig:up3d-ours}
\end{figure}

Our experiment focuses on the type of supervision. Naively training the {\em Priors} networks using an $\mathcal{L}_2$ loss for the $\bm{\theta}$ and $\bm{\beta}$ parameters~\cite{tan2017indirect}, keeps the prediction error high as can be seen in Table~\ref{tab:up3d_ablative} (line~1). Alternatively, we can transform the $\bm{\theta}$ parameters from axis-angle representation to rotation matrix using the Rodrigues' rotation formula~\cite{gallego2015compact}, and apply an $\mathcal{L}_2$ loss on this representation instead (line~2). This leads to more stable training and better performance, as has also been observed by Lassner~\etal~\cite{lassner2017unite}. However, generating the body mesh and further training of the network using our proposed per-vertex supervision (line~3) is even more appropriate and elevates our framework to state-of-the-art performance (see Section~\ref{sec:sota}). Finally, the additional end-to-end finetuning with 2D annotations and the reprojection error (line 4) offers a mild refinement to the network. In the UP-3D case, the benefit is small, since the {\em Priors} have already observed very similar examples with full 3D ground truth, so 2D annotations become redundant. However, when training the {\em Priors} with CMU data, the domain shift, from CMU poses to UP-3D poses is significant, so these 2D annotations offers a clear performance benefit. This is an interesting empirical result demonstrating that training with reprojection losses can be useful not only for end-to-end refinement, but it can also assist the network with novel information recovered from 2D annotations. Some qualitative results from UP-3D using our best model are presented in Figure~\ref{fig:up3d-ours}.

\subsection{Comparison with state-of-the-art}\label{sec:sota}

\noindent
\textbf{UP-3D}:
We compare with two state-of-the-art direct prediction approaches by Lassner~\etal~\cite{lassner2017unite} and Tan~\etal~\cite{tan2017indirect}. We do not include the SMPLify method~\cite{bogo2016keep} since a version of this algorithm was used to generate the ground truth for this dataset, so we observed that many estimated reconstructions had only minimal differences from the ground truth. For~\cite{lassner2017unite} we use the publicly available code to generate predictions. The complete results are presented in Table~\ref{tab:up3d}. Our approach outperforms the other two baselines by significant margins. It is interesting to note that a version of \cite{tan2017indirect}, which uses over 100k images (most of them synthetic) with ground truth pose and shape parameters to directly supervise the network (line `Direct') is outperformed by our approach which does not have access to this data. Finally, in Figure~\ref{fig:up3d-ours}, we provide a qualitative comparison with our closest competitor, the direct prediction approach of~\cite{lassner2017unite}.

\begin{table}
\centering
\small
\hspace{-3mm}
\tabcolsep=0.75mm
\begin{tabular}{@{}lccccc@{}}
\toprule
& LSP & MPII & Fashion & Full  & Reduced \\
\midrule
Lassner~\etal~\cite{lassner2017unite} & 174.4  & 184.3  & 108.0  & 169.8 & 123.6 \\
Tan~\etal~\cite{tan2017indirect} (Indirect)  & -  & -  & -  & - &  189 \\
Tan~\etal~\cite{tan2017indirect} (Direct) & -  & -  & -  & - &  105 \\
Ours  & \bf{127.8} & \bf{110.0} & \bf{106.5}  & \bf{117.7} & \bf{100.5} \\
\bottomrule
\end{tabular}
\vspace{3mm}
\caption{Detailed results on UP-3D~\cite{lassner2017unite}. The numbers are mean per vertex errors (mm), except for the `Reduced' column where only 91 landmarks~\cite{lassner2017unite} contribute to the error. Our approach outperforms the other baselines across the table.}
\label{tab:up3d}
\end{table}

\begin{figure}[t]
	  \centering
	   \includegraphics[width=1\linewidth,trim={0cm 15cm 8.5cm 0cm},clip]{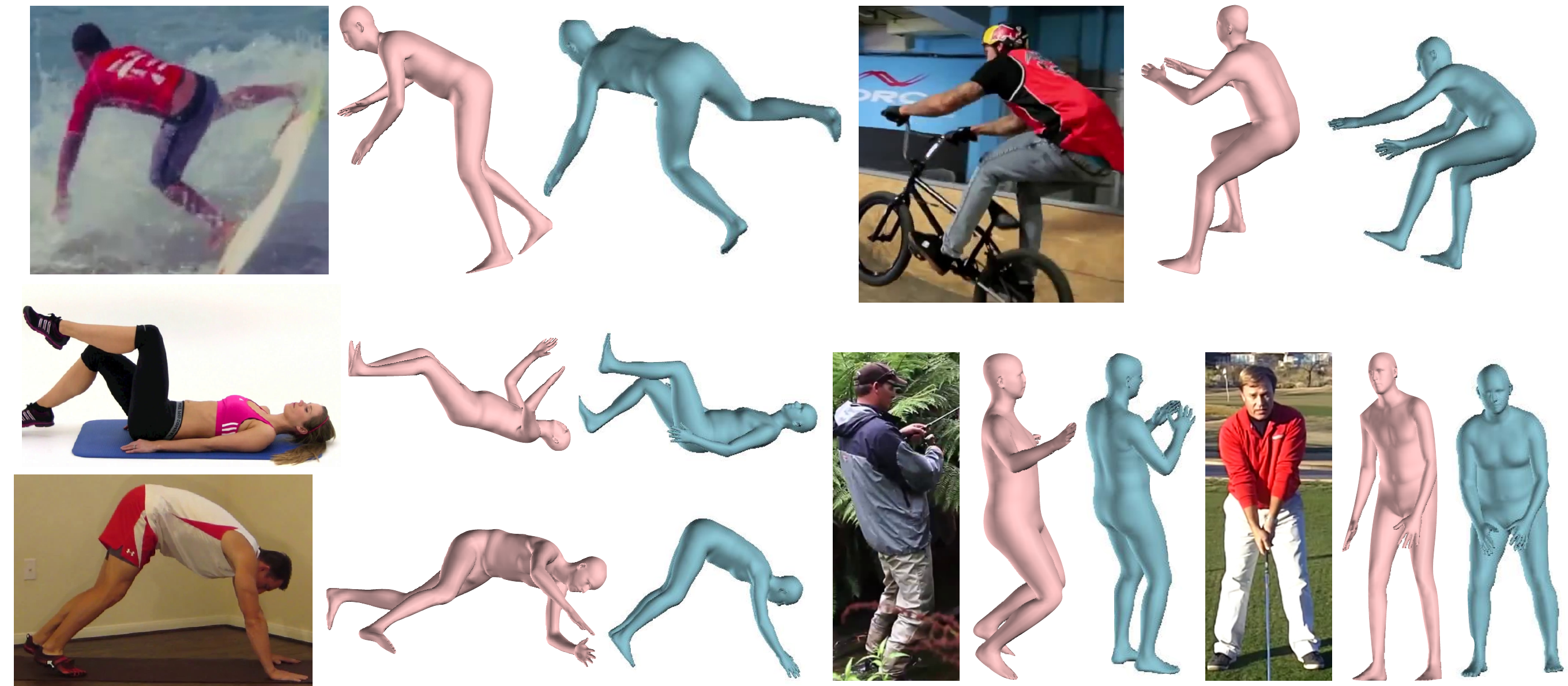}
    \caption{
Examples from UP-3D where our approach (blue shapes) performs significantly better than the direct prediction method of Lassner~\etal~\cite{lassner2017unite} (pink shapes).
}
 \label{fig:up3d-compare}
\end{figure}

\noindent
\textbf{SURREAL}:
We compare with two state-of-the-art approaches, one based on iterative optimization, SMPLify~\cite{bogo2016keep}, and one based on direct prediction~\cite{lassner2017unite}. We use the publicly available code for both approaches to generate predictions. For our approach, we train the {\em PosePrior} using CMU data which we found to be more general than UP-3D. Also, we train two {\em ShapePriors}, for female and male subjects respectively, since the gender is known for this dataset. We emphasize that the testing was conducted on the Human3.6M part of the dataset to avoid any overlap with the training of the different methods (in terms of images or priors). The complete results are presented in Table~\ref{tab:surreal}. Since Lassner~\etal~\cite{lassner2017unite} provide only a non gender-specific model for shape, we also report results considering only the pose estimates, and assuming known shape parameters. Our approach outperforms the other two baselines. For this dataset we observed that because of the challenging color images (low illumination, out-of-context backgrounds, etc), the 2D detections where more noisy than usual, providing some hard failures for the iterative optimization approach~\cite{bogo2016keep}. In contrast, our approach was more resistant to these noisy cases recovering a coherent 3D shape in most cases.

\begin{table}
\centering
\hspace{-3mm}
\tabcolsep=0.75mm
\begin{tabular}{@{}lr@{}}
\toprule
 & Avg\\
\midrule
Lassner~\etal~\cite{lassner2017unite} (GT shape) & 200.5 \\
Bogo~\etal~\cite{bogo2016keep} (GT shape)   & 177.2 \\
Ours (GT shape)  &  {\bf 151.5} \\
\midrule
Bogo~\etal~\cite{bogo2016keep}    & 202.0 \\
Ours  &  {\bf 155.5} \\
\bottomrule
\end{tabular}
\vspace{3mm}
\caption{Detailed results on the Human3.6M part of SURREAL~\cite{varol2017learning}. Numbers are mean per vertex errors (mm). 
``GT shape'' indicates that the shape coefficients are known.
}
\label{tab:surreal}
\end{table}

\begin{table}
\centering
\hspace{-3mm}
\tabcolsep=0.75mm
\begin{tabular}{@{}lr@{}}
\toprule
 & Avg\\
\midrule
Akhter \& Black~\cite{akhter2015pose}* & 181.1\\
Ramakrishna~\etal~\cite{ramakrishna2012}* & 157.3\\
Zhou~\etal~\cite{zhou2016sparse}* & 106.7\\
Bogo~\etal~\cite{bogo2016keep} & 82.3\\
Lassner~\etal~\cite{lassner2017unite} (direct prediction) & 93.9\\
Lassner~\etal~\cite{lassner2017unite} (optimization) & 80.7\\
Ours & \bf{75.9} \\
\bottomrule
\end{tabular}
\vspace{3mm}
\caption{Detailed results on Human3.6M~\cite{ionescu2014human}. Numbers are reconstruction errors (mm). The numbers are taken from the respective papers, except for (*), which were obtained from~\cite{bogo2016keep}.}
\label{tab:hm36m}
\end{table}

\noindent
\textbf{Human3.6M}:
Finally, for Human3.6M we evaluate only the estimated 3D pose, since there is no body shape ground truth available. Our network is the same as before ({\em Priors} trained on CMU), although, we use the 3D joints error for supervision (equation~\ref{eqn:pose3D}), since the focus is on pose. Among others, we compare with the SMPLify method~\cite{bogo2016keep} and the direct prediction approach of Lassner~\etal~\cite{lassner2017unite}. Similarly to the other approaches we compare with, we {\em do not} use any data from this dataset for training. The detailed results are presented in Table~\ref{tab:hm36m}. Our approach again outperforms the other baselines. Some works have reported better results results on Human3.6M (e.g.,~\cite{martinez2017simple,pavlakos2016coarse}), but they do so only by leveraging the training data of this dataset for training.

\subsection{Boosting SMPLify}
In the previous section, we validated that our direct prediction approach can achieve state-of-the-art results with a single prediction step. However, we aspire our method to have greater applicability, by being complementary to iterative optimization solutions. In fact, here we demonstrate that our direct predictions can be a useful initialization and provide a reliable anchor for the SMPLify approach~\cite{bogo2016keep}.

To keep it simple, we make only minor modifications to the SMPLify optimization. First, we use our predicted pose as an initialization, instead of the typical mean pose. Additionally, we avoid the hierarchical four-step optimization, and we limit the whole procedure in a single step. The reason for the multi-stage optimization is to explore the pose space and get a roughly correct pose estimate. However, using our predicted pose as initialization makes this search unnecessary, so we require only the last step of the previously complex optimization scheme. Finally, we add one more data term to the optimization: $E_{anchor}(\bm{\theta}) = \sum_i \rho(\theta_i - \theta^{init}_{i})$, to avoid deviations from our predicted, anchor pose. Similarly to~\cite{bogo2016keep}, we use the Geman-McClure penalty function, $\rho$~\cite{geman1987statistical}, for the optimization. This anchoring, does not typically have effect on the quality of the output, but it can accelerate the convergence. We can also use the shape parameters as anchor, but we observed that pose had greater effect than shape on the optimization.

\begin{table}
\centering
\small
\hspace{-3mm}
\tabcolsep=0.75mm
\begin{tabular}{@{}lcccc@{}}
\toprule
& \multicolumn{2}{c}{FB Seg.} & \multicolumn{2}{c}{Part Seg.} \\
\cmidrule{2-5}
& acc. & f1 & acc. & f1 \\
\midrule
SMPLify & 91.89 & 88.07 & 87.71 & 63.98 \\
SMPLify + our anchor & \bf{92.17} & \bf{88.38} & \bf{88.24} & \bf{64.62} \\
\midrule
SMPLify on GT & 92.17 & 88.23 & 88.82 & 67.03 \\
\bottomrule
\end{tabular}
\vspace{3mm}
\caption{Accuracy and f1 scores for foreground-background and six-part segmentation on LSP test set for different versions of SMPLify. Using our direct prediction as an anchor improves vanilla SMPLify, while also achieving a {\em 3x speedup}. The numbers for the first and third rows are taken from~\cite{lassner2017unite}.
}
\label{tab:dp+smpl}
\end{table}

\begin{figure}[t]
	  \centering
	   \includegraphics[width=1\linewidth,trim={0cm 20cm 0cm 0cm},clip]{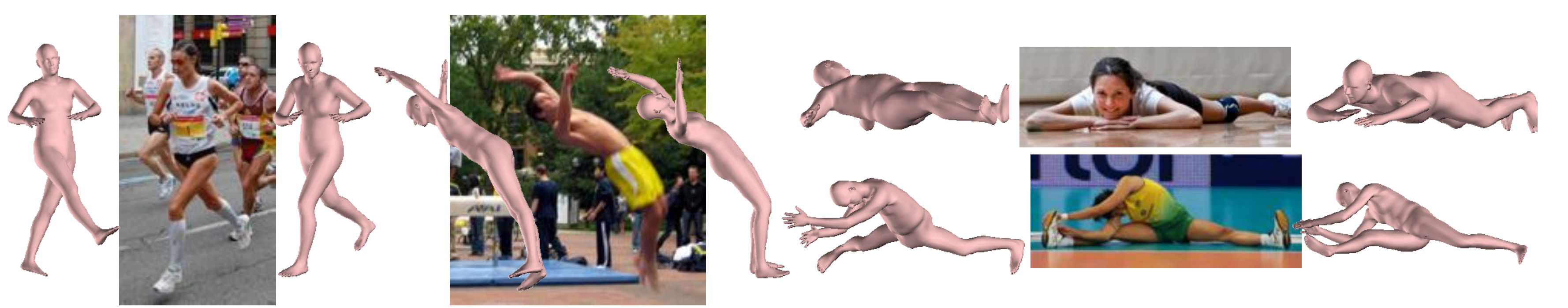}
    \caption{
LSP examples with improved SMPLify fits (right side of each image) when our direct prediction is used as an initialization and anchor for the iterative optimization.
}
 \label{fig:boosting}
\end{figure}

For our evaluation, we use the public implementation of SMPLify and we run the original code, as well as our anchored version, on the LSP test set. The anchored version is {\em three times faster} on average than vanilla SMPLify. More importantly, this speedup comes also with a quantitative performance {\em benefit}. In Table~\ref{tab:dp+smpl} we present the segmentation accuracy of different SMPLify versions, by projecting the 3D shape estimate on the image. To demonstrate that the performance benefit of our anchored version is non-trivial, we report the results for running SMPLify on the ground truth 2D joints and silhouettes. Improved fits from the anchored version are presented in Figure~\ref{tab:dp+smpl}. These results validate the additional benefit of our direct prediction approach, since it can also enhance current pipelines that rely on iterative optimization.

\subsection{Running time}\label{sec:time}

Our approach requires a single forward pass from the ConvNet to estimate the full body 3D human pose and shape. This translates to only 50ms on a Titan X GPU. In comparison, SMPLify~\cite{bogo2016keep} report roughly 1 minute for the optimization, while the publicly available (unoptimized) code runs on 3 minutes per image on average. When the number of landmarks increases to 91, Lassner~\etal~\cite{lassner2017unite} report that the SMPLify optimization can get two times slower. This makes our direct prediction approach more than three orders of magnitude faster than the state-of-the-art iterative optimization approaches. Regarding other direct prediction approaches, Lassner~\etal~\cite{lassner2017unite} reports runtime of 378ms, but we demonstrate significantly better performance 
with our end-to-end framework.

\section{Summary}
The goal of this paper was to present a viable ConvNet-based approach to predict 3D human pose and shape from a single color image. A central part of our solution was the incorporation of a body shape model, SMPL, in the end-to-end framework. Through this inclusion we enabled: a) prediction of the parameters from 2D keypoints and silhouettes, b) generation of the full body 3D mesh at training time using supervision for the surface with a per-vertex loss, and c) integration of a differentiable renderer for further end-to-end refinement using 2D annotations. Our approach achieved state-of-the-art results on relevant benchmarks, outperforming previous direct prediction and optimization-based solutions for 3D pose and shape prediction. Finally, considering the efficiency of our approach, we demonstrated its potential to accelerate {\em and} improve typical iterative optimization pipelines.

\vspace{1em}
\footnotesize
\noindent
{\bf Project Page:} \url{https://www.seas.upenn.edu/~pavlakos/projects/humanshape}

\vspace{0.5em}
\footnotesize
\noindent
{\bf Acknowledgements:} We gratefully appreciate support through the following grants: NSF-IIP-1439681 (I/UCRC), ARL RCTA W911NF-10-2-0016, ONR N00014-17-1-2093, DARPA FLA program and NSF/IUCRC.

{\small
\bibliographystyle{ieee}
\bibliography{egbib}
}

\end{document}